\def\arxiv{}

\ifdefined\arxiv

\documentclass[a4paper,10pt]{article} 

\usepackage{natbib}
\usepackage[utf8]{inputenc}
\usepackage[T1]{fontenc}
\usepackage{geometry}
\usepackage{lipsum} 
\usepackage{parskip} 

\setcitestyle{authoryear,round,citesep={;},aysep={,},yysep={;}}

\usepackage{etoolbox}
\ifundef{\abstract}{}{\patchcmd{\abstract}%
    {\quotation}{\quotation\noindent\ignorespaces}{}{}}

\title{An Emulator for Fine-Tuning Large Language Models \\using Small Language Models}

\else 

\documentclass{article} 
\usepackage{iclr2023_conference}

\usepackage{titlesec}

\titlespacing\section{0pt}{4pt plus 2pt minus 2pt}{2pt plus 2pt minus 2pt}
\titlespacing\subsection{0pt}{4pt plus 4pt minus 2pt}{2pt plus 2pt minus 2pt}

\title{An Emulator for Fine-Tuning Large Language Models using Small Language Models}

\fi

\usepackage{times}

\usepackage{amsmath,amsfonts,bm}

\newcommand{\policy}{\pi}
\newcommand{\reference}{\pi_\text{ref}}

\def\eqref#1{equation~\ref{#1}}

\def\1{\bm{1}}

\DeclareMathAlphabet{\mathsfit}{\encodingdefault}{\sfdefault}{m}{sl}
\SetMathAlphabet{\mathsfit}{bold}{\encodingdefault}{\sfdefault}{bx}{n}

\DeclareMathOperator*{\argmax}{arg\,max}

\usepackage{url}
\usepackage{booktabs}
\usepackage{diagbox}
\usepackage{colortbl}
\usepackage{makecell}
\usepackage{graphicx}
\usepackage{multirow}
\usepackage{wrapfig}

\definecolor{Blue9}{rgb}{0.098,0.3,0.9}
\usepackage[backref=page]{hyperref}
\hypersetup{
  colorlinks,
  urlcolor  = Blue9,
  linkcolor = Blue9,
  citecolor = Blue9,
}
\usepackage[shortcuts]{extdash}

\newcommand{\rev}[1]{{#1}}

\author{\textbf{Eric Mitchell}, \textbf{Rafael Rafailov}, \textbf{Archit Sharma}, \\ \textbf{Chelsea Finn}, \textbf{Christopher D. Manning} \\
Stanford University \\
\texttt{eric.mitchell@cs.stanford.edu}
}

\date{}

\begin{document}

\maketitle

\begin{abstract}
Widely used language models (LMs) are typically built by scaling up a two-stage training pipeline: a pre-training stage that uses a very large, diverse dataset of text and a fine-tuning (sometimes, `alignment') stage that uses targeted examples or other specifications of desired behaviors. While it has been hypothesized that knowledge and skills come from pre-training, and fine-tuning mostly filters this knowledge and skillset, this intuition has not been extensively tested. To aid in doing so, we introduce a novel technique for decoupling the knowledge and skills gained in these two stages, enabling a direct answer to the question, \textit{What would happen if we combined the knowledge learned by a large model during pre-training with the knowledge learned by a small model during fine-tuning (or vice versa)?} Using an RL-based framework derived from recent developments in learning from human preferences, we introduce \textit{emulated fine-tuning (EFT)}, a principled and practical method for sampling from a distribution that approximates \rev{(or `emulates')} the result of pre-training and fine-tuning at different scales. Our experiments with EFT show that scaling up fine-tuning tends to improve helpfulness, while scaling up pre-training tends to improve factuality. Beyond decoupling scale, we show that EFT enables test-time adjustment of competing behavioral traits like helpfulness and harmlessness without additional training. Finally, a special case of emulated fine-tuning, which we call LM \textit{up-scaling}, avoids resource-intensive fine-tuning of large pre-trained models by ensembling them with small fine-tuned models, essentially \rev{emulating} the result of fine-tuning the large pre-trained model. Up-scaling consistently improves helpfulness and factuality of instruction-following models in the Llama, Llama-2, and Falcon families, without additional hyperparameters or training.

\end{abstract}

\section{Introduction}

\begin{figure}
    \centering
    \includegraphics[width=\textwidth]{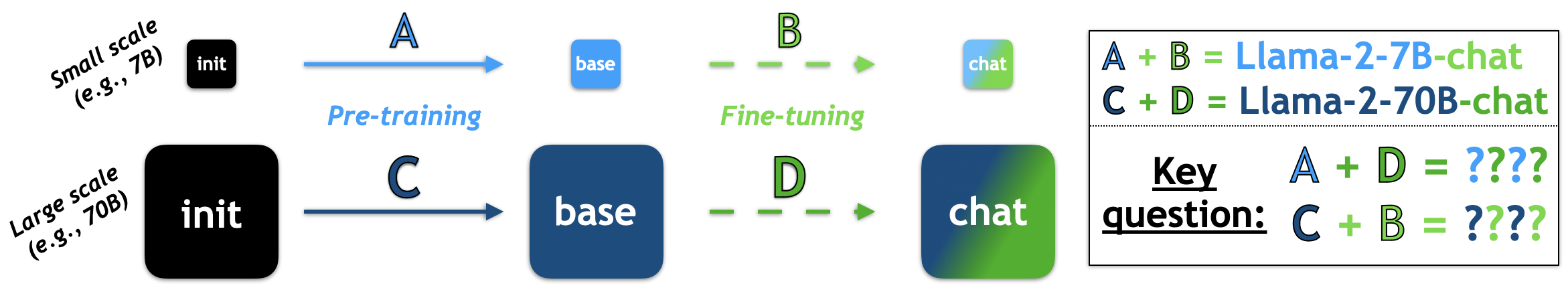}
    \caption{\small \textbf{Emulated fine-tuning (EFT)} enables a principled answer to the question of \textit{what happens when we combine what is learned from pre-training a model of one size with what is learned from fine-tuning a model of a different size?} Conventional models combine the learnings of pre-training and fine-tuning at the same size (A + B, C + D). In contrast, EFT enables choosing these independently, allowing a principled approach to evaluating the result of A + D and C + B.}
    \label{fig:fig1}
\end{figure}

Widely used instruction-following large language models (LLMs) typically follow a two-stage training procedure, with a stage of unsupervised pre-training on a large, diverse dataset followed by supervised fine-tuning on a much smaller, carefully curated dataset \citep{2020t5,chung2022scaling}. While both stages are important in producing models that possess broad world knowledge and perform a given task reliably, identifying exactly what capabilities emerge in which stage and at what scale is difficult \citep{wei2022emergent,schaeffer2023emergent}. For example, pre-trained models typically require careful prompting in order to perform a task; after fine-tuning for instruction following, they typically do not. Evaluation of the extent to which the core capability of `instruction following' is learned during pre-training vs.\ during fine-tuning is thus seriously complicated by the choice of this prompt.
To enable more direct attribution of capabilities to a stage of training, we introduce a principled technique for emulating the result of combining the capabilities gained from pre-training and fine-tuning at different model scales; see Figure~\ref{fig:fig1}. This technique, which we call emulated fine-tuning (EFT), enables\rev{:} a) direct study of the capabilities that change as only one stage is scaled up or down; b) the practical benefit of approximating the result of fine-tuning a large model without the associated computational expense\rev{; and c) the ability to modify the fine-tuning objective (e.g., the tradeoff between helpfulness and harmlessness) at test time, without additional training.}

\begin{figure}
    \centering
    \includegraphics[width=\textwidth]{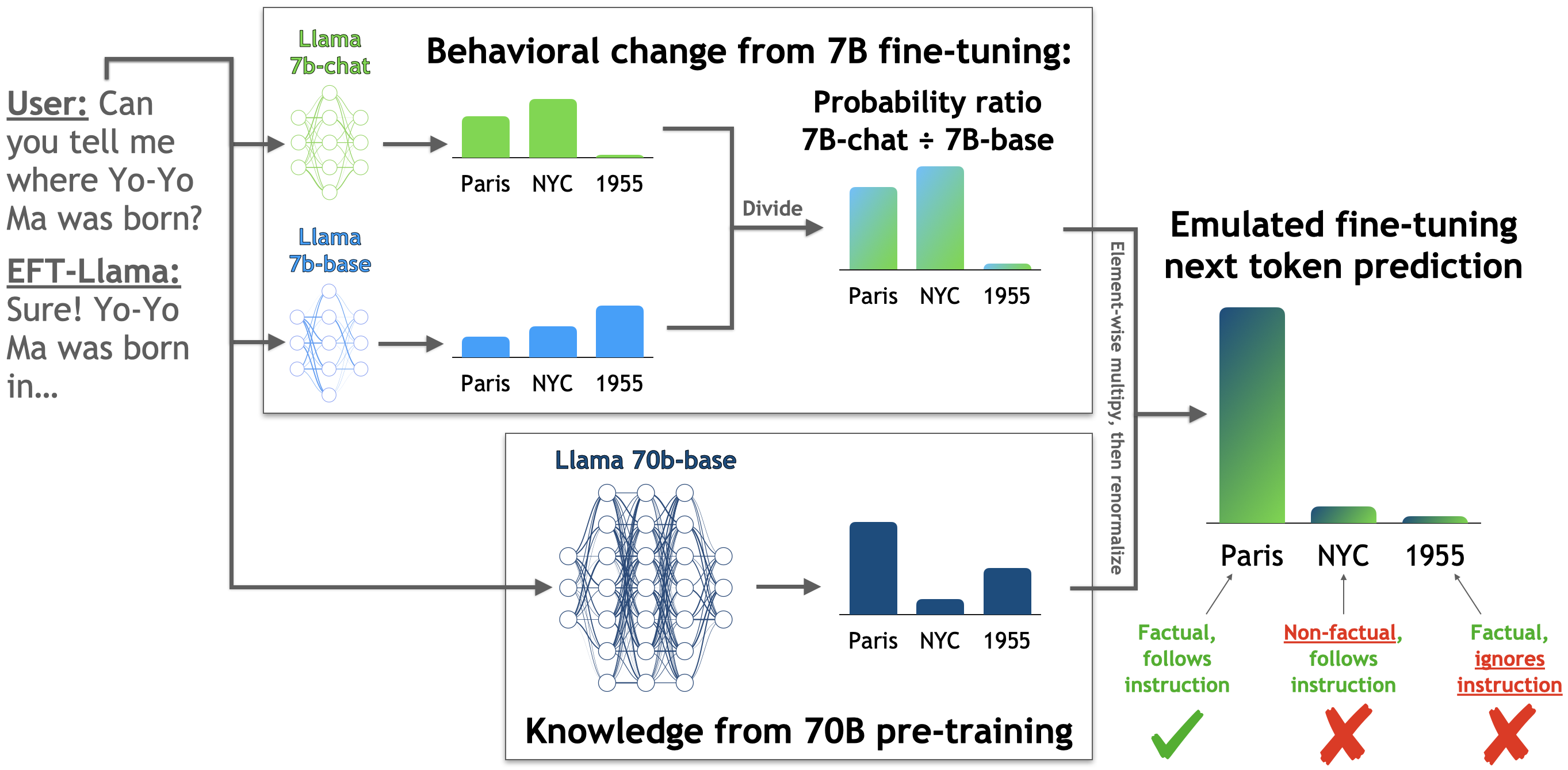}
    \caption{\small \textbf{Emulated fine-tuning combines knowledge from pre-training and fine-tuning at different scales.} This example shows \textit{up-scaling}, which applies the behavioral changes from small-scale fine-tuning to the knowledge in a large pre-trained model. The small fine-tuned model (green) understands the user's query asks about Yo-Yo Ma's place of birth, not year, does not know the correct city. The small pre-trained model (light blue) does not understand the user's query or have reliable knowledge, assigning high probability to the (correct) year of birth of Yo-Yo Ma and both possible places of birth. Their ratio represents the behavior of following user intent (responding only with locations). Reweighting the large base model's \textit{factually correct} conditional (that fails to follow user intent) using the small-scale behavioral change ratio, we emulate what a large scale fine-tuned model \textit{would have said}: a factually correct response that also follows the user's intent.}
    \label{fig:method}
\end{figure}

Emulated fine-tuning is based on a simple factorization of the logits of a fine-tuned language model into a) the base log probabilities of a pre-trained base model and b) the `behavior delta', or the difference between the log probabilities of a base model and fine-tuned model. This delta is a compact representation of the behavior change learned in fine-tuning and can be justified through either a reinforcement learning \citep{rafailov2023direct} or Bayesian inference \citep{korbak-etal-2022-rl} framework. EFT thus emulates the result of pre-training at one scale and fine-tuning at another by adding base log probabilities computed by a model at one size and the behavior delta computed by models of a different size. For example, using models from the Llama-2 family, we can emulate the result of pre-training at 70B scale and fine-tuning at 7B scale by performing the log probability algebra \textbf{Llama-2-base 70B + (Llama-2-chat 7B - Llama-2-base 7B)}, where the first term is the base log probabilities and the term in parentheses is the behavioral delta. See Figure~\ref{fig:method} for a concrete example of this emulation.

Using emulated fine-tuning, we analyze the results of pre-training and fine-tuning at various scales for multiple model families and datasets. Our analyses provide evidence supporting the intuition that pre-training at scale enables greater accumulation of raw knowledge (improved factual correctness), while fine-tuning at larger scale produces greater helpfulness (improved user satisfaction) \citep[cf.][]{gudibande2023false}. Beyond this scientific finding, we also find that EFT enables boosting the performance of small fine-tuned models by a process we call \textit{up-scaling}, essentially ensembling the small fine-tuned model with a larger pre-trained model, without any fine-tuning or modifications to either model. Our experiments show that in scenarios where fine-tuning a small language model is viable (e.g., Falcon-7B) but fine-tuning a larger language model is not due to resource constraints (e.g., Falcon-180B), up-scaling enables capturing much of the benefits of fine-tuning the larger model, without performing any model fine-tuning. \rev{Finally, we show that EFT also enables emulating modifications the fine-tuning objective at test time through the mixing of different behavioral deltas with different weightings.}

In summary, our primary contributions are a) the emulated fine-tuning framework; b) clear experimental justification for the claim that scaling pre-training leads to improved factual knowledge while scaling fine-tuning leads to improved task adherence; and c) the technique of model \textit{up-scaling}, which enables a small fine-tuned model and large base model to approximate the compute-intensive result of fine-tuning a large base model.

\section{Related Work}
The benefits of unsupervised pre-training in neural networks was first identified in deep belief networks \citep{hinton2006fast} and stacked autoencoders \citep{bengio2007greedy}, with early analyses noting persistent effects of pre-training even when fine-tuning data is not limited \citep{erhan2010unsupervised}. In natural language processing, pre-trained representations of individual words \citep{word2vec,pennington-etal-2014-glove} or entire passages \citep{devlin-etal-2019-bert,peters-etal-2018-deep} demonstrated the ability for task-agnostic pre-training to learn representations useful for a wide variety of downstream linguistic tasks such as question-answering, natural language inference, and translation \citep{devlin-etal-2019-bert,2020t5}. The transformer architecture \citep{NIPS2017_3f5ee243} enabled more efficient pre-training on large datasets, which proved to inject significant amounts of precise factual world knowledge into pre-trained LMs \citep{petroni-etal-2019-language} that can be redirected to downstream tasks through fine-tuning \citep{roberts-etal-2020-much}. Most recently, various works have shown that language models pre-trained with unsupervised generative modeling can be fine-tuned to engage in general-purpose dialogue, producing a model that can perform a variety of complex tasks specified in natural language \citep{thoppilan2022lamda,ouyang2022training,bai2022training,bubeck2023sparks,touvron2023llama2}. Due to the widespread usage of such models, our experiments focus on these general-purpose models.

Increasing model scale has proven a key aspect of increasing the benefits of pre-training to fluency, world knowledge, reasoning ability, and a variety of other properties \citep{gpt3,kaplan2020scaling,touvron2023llama}. Other work leverages this capability differential to improve language model sampling through `contrastive decoding', subtracting the log probabilities of a small language model (scaled by a small constant hyperparameter) from the log probabilities of a large language model \citep{li-etal-2023-contrastive}. Our work differs by interpreting this log probability difference as a log-importance weight, using it to re-weight the log probabilities of another model and eliminating the need \rev{for the added scaling hyperparameter}. Relatedly, \citet{gao2022scaling} study the impact of scale on the reward model used during RLHF, which can be interpreted as scaling the fine-tuning phase in our work; however, they do not explore pre-training scale or investigate the impact of either scale on independent model capabilities. \rev{In concurrent work, \citet{deng2023rewardaugmented} train a reward model that reweights a base model's conditional distributions during sampling. Our work differs in that EFT does not require training a new reward model, has a principled basis in reinforcement learning, and scales more efficiently with the vocabulary size, due to the parameterization of the reward as a ratio of log probabilities \citep{rafailov2023direct}.}

\section{Emulated Fine-Tuning: Decoupling the Scale of Pre-training and Fine-tuning}

We now describe the framework of emulated fine-tuning (EFT) and how it enables decoupling the scale of pre-training and fine-tuning, as well as \textit{up-scaling}, a special case of emulated fine-tuning that is particularly useful in practice.

\subsection{Preliminaries}
Emulated fine-tuning views the fine-tuning procedure as reinforcement learning (RL) with a KL-divergence constraint preventing divergence from a reference model, in this case the pre-trained model \citep{peters2010relative}. That is, we view the result of fine-tuning $\pi_\text{ft}$ as the solution to
\begin{align}
    \policy_\text{ft} = \policy^*(r,\reference) = \argmax_\policy\;&\mathbb{E}_{x\sim p(x),y\sim \policy(\cdot \mid x)} \left[r(x,y) - \beta \text{KL}(\policy(\cdot \mid x) \| \reference(\cdot \mid x))\right]
    \label{eq:kl-rl}
\end{align}
where $\beta$ controls the strength of the KL constraint to the pre-trained model (the reference model) \rev{and $p(x)$ is a fixed distribution (or dataset) of prompts}. Prior work \citep{peters2010relative,peng2019advantageweighted,korbak-etal-2022-rl,rafailov2023direct} shows that the solution is given by 
\begin{equation}
    \pi^*(r,\reference)(y \mid x) = \frac{1}{Z(x)} \pi_\text{ref}(y\mid x)\exp\left(\frac{1}{\beta} r(x,y)\right),
\end{equation}
with $Z(x) = \sum_y \pi_\text{ref}(y \mid x)\exp\left(\frac{1}{\beta} r(x,y)\right)$. 
Crucially, while the EFT framework is justified with an RL-based interpretation of fine-tuning, it is applicable to \textit{any} fine-tuned model, as any language model can be viewed as the solution to KL-constrained RL with a constraint to the pre-trained model \citep{rafailov2023direct}. Specifically, any fine-tuned language model $\pi_\text{ft}$ and pre-trained model $\reference$ can be mapped to a reward function $r_{\policy_\text{ft}}(x,y)$ such that the solution to the KL-constrained RL problem $\policy^*(r_{\policy_\text{ft}},\reference) = \pi_\text{ft}$, using $r_{\policy_\text{ft}}(x,y) = \beta \log \frac{\pi_\text{ft}(y \mid x)}{\reference(y \mid x)}$.

Using this duality between language models and rewards, for any language model $\policy_\text{ft}$ fine-tuned from a pre-trained model $\reference$, we can re-write
\begin{align}
    \policy_\text{ft}(y \mid x) = \reference(y \mid x) \exp\biggl(\underbrace{\log \frac{{\policy_\text{ft}}(y \mid x)}{\reference(y \mid x)}}_\text{Implicit reward}\biggr) = \reference(y \mid x) \exp\biggl(r_{\policy_\text{ft}}(x, y)\biggr)
    \label{eq:eft}
\end{align}
In other words, the fine-tuned model $\policy_\text{ft}$ is the optimal policy to the KL-constrained reward maximization problem with reward function $r_{\policy_\text{ft}}(x,y) = \log \frac{{\policy_\text{ft}}(y \mid x)}{\reference(y \mid x)}$, using $\reference$ as the reference model that we are constraining to. We now have a clear delineation of the loci of information gained from pre-training and fine-tuning: pre-training knowledge is represented in the base log probabilities, while capabilities gained from fine-tuning are captured in the reward (the behavior delta of base log probabilities subtracted from fine-tuned model log probabilities). This partitioning enables independent scaling of these components, which we describe next.

\subsection{Scale Decoupling with EFT}
\label{sec:scale-decouple}
To make explicit the size of model used to compute the corresponding conditionals, we add superscripts \rev{and subscripts to Eq.~\ref{eq:eft} denoting the scale of the model used to compute each quantity}:
\begin{equation}
    \policy^N_M(y \mid x) = \frac{1}{Z^N_M(x)}\reference^N(y \mid x) \exp\Bigl(r^M_\policy(x, y)\Bigr) \propto\reference^N(y \mid x)\frac{\policy^M(y\mid x)}{\reference^M(y \mid x)}
    \label{eq:decoupled-eft}
\end{equation}
where the $M$-scale reward function is $r^M_\policy(x,y) = \log \frac{\policy^M(y \mid x)}{\reference^M(y \mid x)}$ and the scale-decoupled partition function is $Z^N_M(x) = \sum_y \pi_\text{ref}^N(y \mid x)\exp\left(r^M(x,y)\right)$.\footnote{The partition function appears here, but not Eq~\ref{eq:eft}, as the reference models are no longer exactly equal (they are different sizes).} That is, $\policy^N_M$ corresponds to simulating mixing the knowledge learned by a model of size $N$ during pre-training and the knowledge learned by a model of size $M$ during fine-tuning. While setting $N=M$ corresponds to simply sampling from the original policy, in this paper, we particularly explore the setting of $N \neq M$. For $N < M$, we simulate mixing the knowledge of a small reference (pre-trained) model with the knowledge learned by a \textit{large} model during fine-tuning; for $N > M$, we simulate mixing the knowledge of a large pre-trained model with the knowledge learned by a \textit{small} model during fine-tuning.

\textbf{Sampling with Emulated Fine-tuning.} Our experiments rely on drawing samples from EFT models. To do so, we compute per-token conditionals according to Eq.~\ref{eq:decoupled-eft}, but use a per-timestep approximation of the (intractable) sequence-level partition function:
\begin{equation}
    \tilde \pi(y_t \mid x, y_{<t}) = \frac{1}{Z(x,y_{<t})}\reference^N(y_t \mid x, y_{<t}) \frac{\pi^M(y_t \mid x, y_{<t})}{\reference^M(y_t \mid x, y_{<t})},
\end{equation}
with per-timestep partition function $Z(x,y_{<t})=\sum_{y_t} \reference^N(y_t \mid x, y_{<t}) \frac{\pi^M(y_t \mid x, y_{<t})}{\reference^M(y_t \mid x, y_{<t})}$. A similar \rev{temporally} greedy approximation emerges from recent work in preference learning that interprets preference learning not as learning a \textit{reward function}, but rather an \textit{advantage function} \citep{knox2023models}.

\subsection{Computational Factors and Language Model Up-Scaling}
Emulated fine-tuning enables sampling from an approximation of the result of pre-training and fine-tuning at different scales. We refer to the case when $N>M$ as \textit{up-scaling}, as we emulate the result of fine-tuning \rev{a \textit{large} model}; we refer to the case of $N<M$ as \textit{down-scaling}, as we emulate the result of fine-tuning \rev{a \textit{small} model}. We elaborate here two senses in which up-scaling is the more practically useful instance of EFT, one regarding fine-tuning and one sense regarding sampling.

\begin{wrapfigure}{r}{0.42\textwidth}
        \centering
        \vspace{-8mm}
        \includegraphics[width=0.42\textwidth]{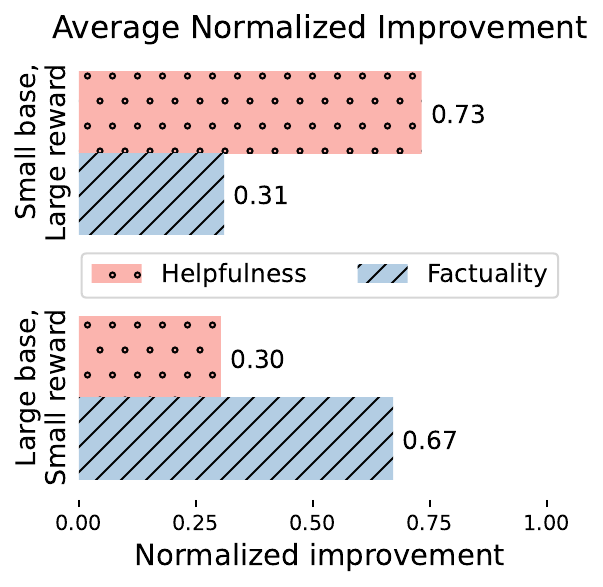}
        \vspace{-5mm}
        \caption{\small \textbf{Scaling pre-training alone mostly benefits factuality; scaling up fine-tuning alone mostly benefits helpfulness.} The bottom group of bars shows that emulating a large fine-tuned model with a small fine-tuned model and large base model produces nearly 70\% of the factuality gains compared to the small fine-tuned model alone. Normalized improvements averaged across Llama-1, Llama-2, and Falcon model families and Anthropic-HH and ELI5 datasets.}
        \vspace{-4mm}
        \label{fig:averages}
\end{wrapfigure}First, down-scaling assumes access to the \textit{actual} fine-tuned model at the larger scale, in order to simulate the result of fine-tuning at smaller scale. In this case, simply sampling from the large fine-tuned model would be computationally cheaper and more efficient. In contrast, up-scaling assumes access to a small fine-tuned model \rev{for the specific task or domain of interest} (computationally cheap to acquire) and a large pre-trained model (many of which are freely released by organizations with considerable resources). Second, sampling from an EFT model with $N \gg M$ is more efficient: EFT sampling requires computing one forward pass of a model at size $N$ (the $N$-scale pre-trained model) and \textit{two} forward passes of models at size $M$ (the $N$-scale fine-tuned model and the $N$-scale pre-trained model). As $N$ becomes much larger than $M$, this computational cost becomes \rev{essentially the same as} sampling from the actual $N$-scale fine-tuned model. Further, if $M$ is small relative to $N$, a natural adaptation of speculative decoding \citep{leviathan2023fast,chen2023accelerating} to EFT exists, in which the $M$-scale fine-tuned model proposes chunks of tokens for the full EFT model to check. Section~\ref{sec:speculative-decoding} confirms that speculative decoding can enables a \rev{nearly 2.5x} speedup for sampling from up-scaled models, without changing the model's samples.

For these reasons, EFT up-scaling is a more practically useful technique to improving the performance of small, fine-tuned language models.

\section{Experiments}
\label{sec:experiments}

Our experiments primarily address the question \textit{what capabilities change when independently scaling pre-training vs fine-tuning?} \rev{To answer this question, we} use EFT to evaluate helpfulness and factuality of a variety of scale combinations. \rev{We also attempt interpolating between different behavior deltas with EFT, for example to change the desired tradeoff between helpfulness and harmlessness at test time, without additional training.} Next, we show that up-scaling with EFT requires modifying the small fine-tuned model's conditional for a sparse set of timesteps, enabling a large speedup in sampling by adapting speculative decoding to EFT up-scaling. We also conduct an ablation to show some potential benefits of filtering noisy token reweightings. Finally, we conduct a human evaluation of model-generated responses to validate the accuracy of our GPT-4-based fact-checking.

\paragraph{Datasets}
Our experiments use two datasets that assess a dialogue agent's ability to provide helpful, factual assistance to a user. First, we use the \textbf{Anthropic Helpful-Harmless (HH)} dialogue dataset \citep{bai2022training}, which consists of multi-turn dialogue between a human and chatbot. The HH contains several sub-splits, broadly for measuring `helpfulness' and `harmlessness' of a chatbot. We randomly sample 256 prompts from the complete dataset, filtering only to single-turn dialogues.\footnote{This choice is to prevent GPT-4 evaluating responses in the dialogue history that didn't come from the EFT model during evaluation.} Second, we use prompts from the ELI5 \citep{fan-etal-2019-eli5} dataset, a dataset of open-ended user-generated questions about science, history, and everyday life sourced from the Reddit ELI5 forum.  We select a random subset of 256 ELI5 prompts from test split, filtering to queries with no more than 30 words. Prompts in the HH dataset are more everyday and conversational, asking for movie recommendations or instructions for home maintanence tasks. In contrast, ELI5 prompts tend to \rev{ask} more difficult, targeted factual questions about scientific or political topics.

\paragraph{Models.} Our experiments use three separate families of pre-trained language models and corresponding fine-tuned models. For our \textbf{Llama-1} experiments, we use the Llama-1 base models \citep{touvron2023llama} at 7B and 65B scale and Vicuna fine-tuned models \citep{vicuna2023} at 7B and 33B scale (no 70B Vicuna model is available) to compute implicit rewards. Vicuna models are fine-tuned from Llama-1 base models with on publicly-shared conversations that users have had with ChatGPT. Our \textbf{Llama-2} experiments use the Llama-2 base models \citep{touvron2023llama2} at 7B and 70B scale and Llama-2-chat models at 7B and 70B scale to compute implicit rewards. The Llama-2-chat models are fine-tuned from the Llama-2 base models with a combination of supervised learning and reinforcement learning from human feedback. Finally, for our \textbf{Falcon} experiments, we use Falcon base models \citep{falcon40b} at 7B and 180B scale and the Falcon instruct/chat models at 7B and 180B scale to compute implicit rewards.\footnote{Due to GPU memory constraints, we use Falcon-180B in 8bit inference mode when computing large-scale rewards for the Falcon down-scaling experiments, as both the 180B chat and base models cannot fit on 8 A100s in float16; quantization is likely to have some effect on generation quality. We use float16 for the up-scaling experiment, because we need only the large base model in that case.} Similarly to Vicuna, Falcon instruct/chat models are fine-tuned with supervised learning on shared dialogues between humans and chatbots. All three families include base generative models pre-trained with unsupervised pre-training on very large, diverse datasets of internet text \citep{touvron2023llama,touvron2023llama2,falcon40b}.

\paragraph{Evaluation.} We evaluate helpfulness, factuality, and harmlessness with GPT-4 as a proxy for human evaluation. Several existing studies have demonstrated the effectiveness of both pair-wise evaluation (comparing the quality of two responses) and point-wise evaluation (scoring a single response along some dimension) using ChatGPT or GPT-4 \citep{zheng2023judging,dubois2023alpacafarm,rafailov2023direct,chen2023exploring} as well as these models' ability to provide well-calibrated judgments of truthfulness \citep{tian2023just}. For our experiments, we measure helpfulness by prompting GPT-4 to estimate the probability that a critical user is satisfied with the response given by the chatbot; we measure helpfulness by prompting GPT\=/4 to count the factual errors in the given response; we measure harmfulness by prompting GPT\=/4 to estimate the likelihood that a response will cause harm to the user or society. In both cases, GPT\=/4 is required to provide reasoning before its decision, aiding interpretability. We sample responses with temperature 0. Further, we conduct a comparison with crowd-sourced annotators in Section~\ref{sec:human-eval}, finding that in the cases of disagreements between GPT-4 and humans, errors in the human judgment, rather than GPT-4's analysis, cause the disagreement nearly 80\% of the time. Complete prompts for GPT-4 evaluations can be found in Appendix~\ref{sec:prompts}.

\begin{figure}
    \centering
    \includegraphics[width=0.9\textwidth]{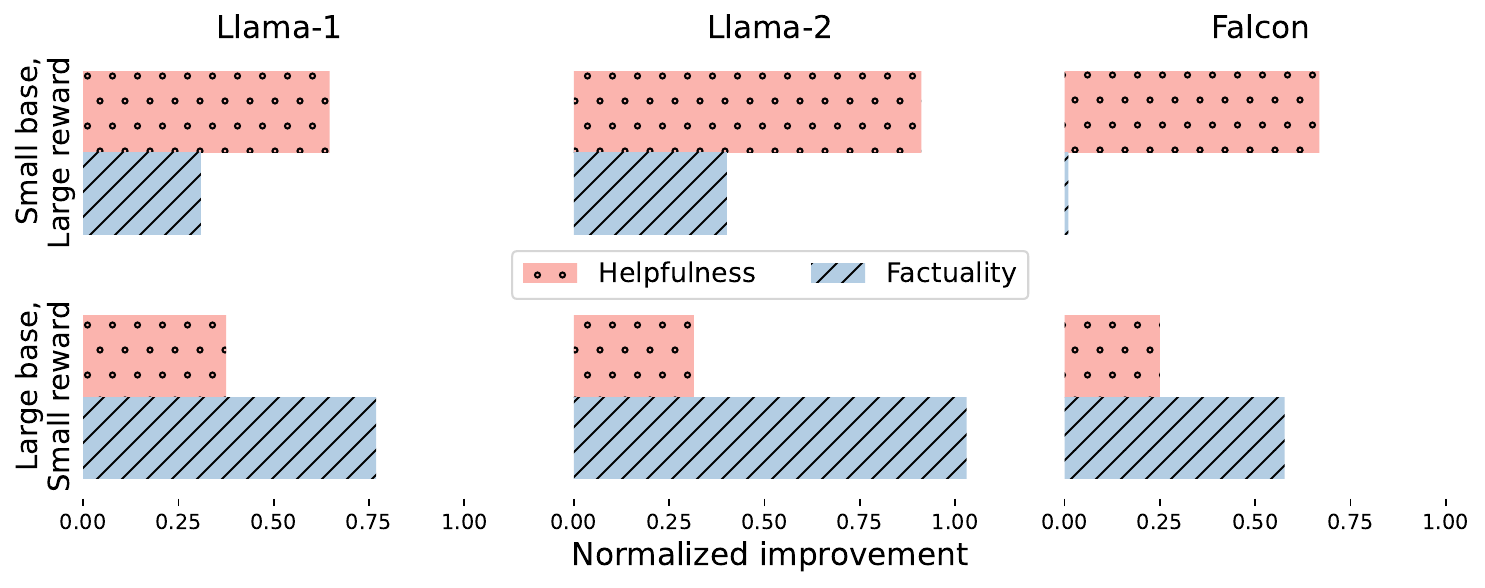}
    \caption{\small Normalized improvements in \textbf{factuality} and \textbf{helpfulness} from emulated fine-tuning for prompts from Anthropic-HH dialogue dataset. Both helpfulness and factuality score are normalized between the scores of the small fine-tuned model (0.0) and the large fine-tuned model (1.0). Up-scaling (bottom row) combines the behavioral adjustments from fine-tuning at small scale with the knowledge gained by pre-training at large scale, and tends to provide more improvement in factuality. Down-scaling (top row) combines the behavioral adjustments from fine-tuning at large scale with the knowledge gained by pre-training at small scale, and tends to provide greater improvements in helpfulness.}
    \label{fig:anthropic-main}
\end{figure}


\subsection{What Capabilities Arise from Scaling Pre-training vs Fine-tuning?}
Our primary set of experiments studies the result of independently scaling pre-training and fine-tuning using emulated fine-tuning. For each dataset and model family, we generate responses to all 256 evaluation prompts using four models: a) the small fine-tuned model alone; b) the large fine-tuned model \rev{alone}; c) the EFT \textit{up-scaled} model, emulating the combination of small-scale fine-tuning and large-scale pre-trained knowledge; d) the EFT \textit{down-scaled} model, emulating the combination of large-scale fine-tuning with small-scale pre-trained knowledge. For example, for the Llama-2 experiments, we sample from a) Llama-2-chat 7B; b) Llama-2-chat 70B; c) up-scaled EFT with Llama-2-base 70B as the pre-trained model and Llama-2-chat 7B/Llama-2-base 7B as the implicit reward; and c) down-scaled EFT with Llama-2-base 7B as the pre-trained model and Llama-2-chat 70B/Llama-2-base 70B as the implicit reward. All experiments use temperature sampling with temperature 1.0, \rev{without top-p or top-k (except when specified otherwise)}.

See Figure~\ref{fig:averages} for the aggregated results of this experiment, which shows evidence that \rev{scaling pre-training primarily leads to improved factuality, while scaling fine-tuning primarily leads to improved perceived helpfulness}. See Figures~\ref{fig:anthropic-main} and~\ref{fig:eli5} for the per-model and per-dataset results. Results are normalized against the performance of the small and large fine-tuned models alone (which are essentially lower and upper bounds on performance); a value of 0.0 corresponds to small fine-tuned model performance, while a value of 1.0 corresponds to large fine-tuned model performance. Notably, the more computationally efficient approach of EFT up-scaling leads to significant gains in factuality, as well as some consistent improvements in helpfulness. Section~\ref{sec:speculative-decoding} explores an approach to making decoding from EFT up-scaled models more efficient.

\subsection{EFT Enables Dynamic Test-Time Reward Interpolation}
\begin{wrapfigure}{r}{0.55\textwidth}
    \vspace{-6mm}
    \centering
    \includegraphics[width=0.5\textwidth]{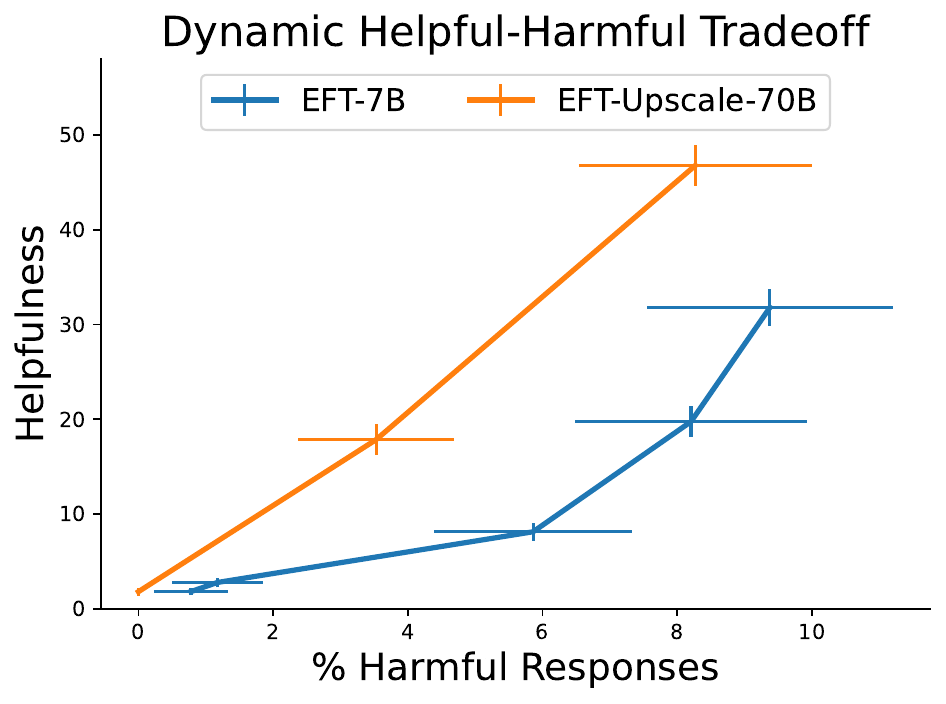}
    \label{fig:tradeoff}
    \vspace{-3mm}
    \caption{\small \small\textbf{Dynamically adjusting the desired tradeoff between helpfulness and harmlessness without retraining}. We use EFT to interpolate between two implicit rewards for helpfulness and harmlessness and plot GPT-4-evaluated helpfulness and fraction of responses that are harmful on Anthropic-HH prompts. Combining reward interpolation with up-scaling enables a Pareto improvement in the frontier, \textbf{all without fine-tuning}. Error bars are one standard error.}
    \vspace{-5mm}
\end{wrapfigure}While decoupling scale is a clear feature of EFT, another benefit of explicitly decoupled pre-training and fine-tuning is the ability to make modifications to the reward function at sampling time. Consider the case of competing fine-tuning objectives, such as the objectives of helpfulness and harmlessness \citep{bai2022training}; some user queries (`How can I steal my neighbor's guitars?'), providing an answer that helps the user with their goal is directly at odds with providing a harmless (or safe) answer. Thus, one view of fine-tuning general dialogue agents is attempting to provide maximum helpfulness at a particular budget of harmfulness. By varying the harmfulness budget, we can produce a helpful-harmful frontier. However, existing fine-tuning procedures \textit{bake in} the particular desired tradeoff between helpfulness and harmfulness \rev{at fine-tuning time, and this tradeoff} cannot be easily modified at sampling time.

In contrast, with emulated fine-tuning, such test-time modulation of the reward is natural and straightforward. Figure 5 shows the results of interpolating between helpfulness and harmlessness at 7B pre-training and fine-tuning scale, as well as with up-scaling the pre-trained model to 70B. We see clear, smooth frontiers, and up-scaling provides a Pareto improvement, \textbf{all without retraining to each tradeoff}.

\rev{To interpolate behaviors at test time with EFT, we} assume that two small-scale fine-tuned models exist, one fine-tuned for pure helpfulness $\pi_\text{help}$, one for pure harmlessness $\pi_\text{safe}$. \rev{For this experiment, we} fine-tune these two models with DPO using Llama-2-7B as the base model, and the helpful-base and harmless-base splits of the Anthropic-HH dataset \citep{bai2022training}. At test time, instead of using a single reward function $r_\pi^M(x,y)$ in Equation~\ref{eq:decoupled-eft}, we use the interpolated reward $r_\lambda^M(x,y) = \lambda r_\text{help}^M(x,y) + (1-\lambda) \pi_\text{safe}^M$, where $\lambda=1$ corresponds to pure helpfulness, and $\lambda=0$ pure harmlessness. Sampling with $\lambda \in (0,1)$ corresponds to some mixture of helpful and harmless. We can also combine reward interpolation with model up-scaling in order to \textit{emulate fine-tuning a large pre-trained model with some mixtures of reward functions.}

\subsection{Efficient Sampling from Up-scaled Models with Speculative Decoding}
\label{sec:speculative-decoding}
Naively, EFT up-scaling (small-scale fine-tuning + large pre-trained model) requires two forward passes from the `small' models and one forward pass from the `large' model for each token. Yet the size asymmetry of EFT makes speculative decoding \citep{chen2023accelerating} a natural choice to accelerate inference. Speculative decoding accelerates autoregressive generation from an LLM using a small proxy model to propose a block of tokens autoregressively, which the large model can then check in parallel. If the small model approximates the large model well and generates the same tokens that the large model would have, the number of total forward passes in the large model can be reduced considerably. For EFT up-scaling, we hypothesize that the small fine-tuned model alone might approximate the up-scaled model for most tokens; we verify this hypothesis qualitatively in Figure~\ref{fig:speculative}, which shows that the total variation distance between the small fine-tuned model and the up-scaled model is small for most tokens\rev{, and very large for a few tokens}. Thus, speculative decoding is likely to accelerate EFT up-scaling.

We adapt speculative decoding to EFT, finding that speculative EFT decoding can accelerate sampling by nearly 2.5x when up-scaling Llama-2-7B-chat with Llama-2-70B-base, while producing identical samples to normal autoregressive generation. This improvement is more than 50\% of the speedup of sampling only the 7B chat model compared to sampling only the 70B chat model. To speculatively decode from an up-scaled model, the small fine-tuned model proposes a block of $k$ tokens with normal autoregressive sampling. Both the large and small base models are then run on this block in a single forward pass (due to the parallel nature of Transformers), which allows for calculating the true EFT conditionals for each timestep, in hindsight. If sampling from the true conditionals produces the same tokens\footnote{We set the random seed to be equal to the timestep, to ensure high-entropy conditionals are not penalized.}, we simply continue and sample a new proposed block. In the case of a disagreement, we rewind generation back to the last token \rev{where the small fine-tuned model and complete up-scaled model agreed}. If no tokens agree, we use the token sampled from the first true hindsight up-scaled conditional.

\begin{figure}
    \centering
    \includegraphics[width=0.9\textwidth]{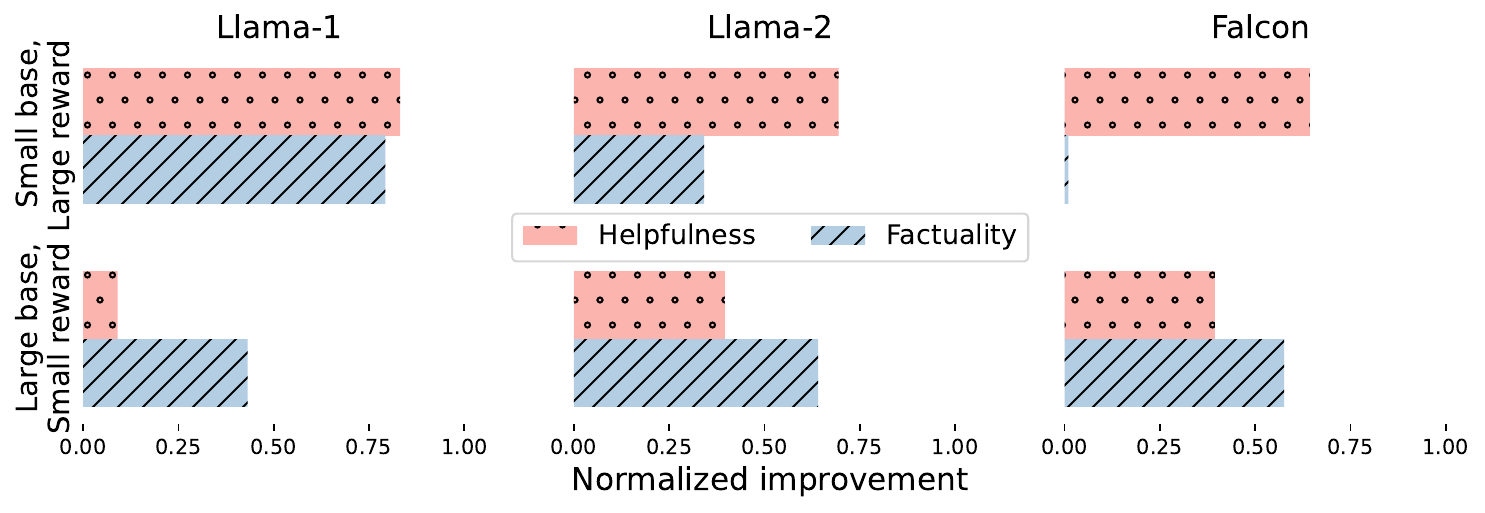}
    \caption{\small Normalized improvements in \textbf{factuality} and \textbf{helpfulness} from emulated fine-tuning on prompts from ELI5 dataset. Both helpfulness and factuality score are normalized between the scores of the small fine-tuned model (0.0) and the large fine-tuned model (1.0).  Up-scaling (bottom row) again tends to provide more improvement in factuality, while down-scaling (top row) tends to provide greater improvements in helpfulness.}
    \label{fig:eli5}
\end{figure}

\subsection{Conservative Decoding Strategies for Up-Scaled Models}
\begin{wraptable}{r}{0.55\textwidth}
    \small
    \centering
    \vspace{-4mm}
    \begin{tabular}{lcccc}
        \toprule
        \textbf{Truncation} & None & 0.95 & 0.9 & 0.8 \\
        \midrule
        \textbf{Errors/prompt} & 0.300 & \textbf{0.289} & 0.352 & 0.348 \\
        \textbf{Helpfulness} & 66.8 & 67.0 & \textbf{67.2} & 67.0 \\
        \bottomrule
    \end{tabular}
    \caption{\small Evaluating \textbf{conservative re-weighting} in up-scaled Llama-2 models by truncating up-scaling weights for low-probability tokens. Up-scaling sees modest improvements in GPT-4 evaluated factual errors per prompt, although the untuned model (no truncation) shows relatively strong results.}
    \vspace{-1mm}
    \label{tab:conservative}
\end{wraptable}All of our prior experiments simply sample from the raw re-weighted conditionals described in Equation~\ref{eq:decoupled-eft}, without introducing any new decoding strategies or hyperparameters. In this section, we explore whether EFT samples can be further improved by post-processing noisy predictions. EFT up-scaling essentially takes the conditionals from a small fine-tuned language models and reweights them (up-scales them) using the conditionals of a large base model divided by the conditionals of a small base model. However, the up-scaling ratio $\frac{p_\text{base-large}(x_t\mid x_{<t})}{p_\text{base-small}(x_t\mid x_{<t})}$ may become extremely large for low-probability (and possibly poorly-modeled) tokens, leading to problematically high probability assigned to low-quality tokens.

 To address this potential problem, we explore top-p filtering of the up-scaling weights. See Table~\ref{tab:conservative} for complete results, showing that top-p filtering of up-scaling weights produces mild improvements in factuality and helpfulness compared to sampling from the unfiltered conditionals. To perform top-p filtering, we first compute the `top-p' set of tokens from the conditional of only the small fine-tuned model, that is, the smallest set of tokens whose probability sums to over $p$. However, unlike conventional top-p decoding \citep{Holtzman2020The}, we do not set the conditionals to other tokens to zero. Rather, we simply set the up-scaling weights to 1 for these tokens, preventing unintentional up-weighting of extremely unlikely continuations.

\subsection{Comparing GPT-4 Factuality Judgments with Human Evaluators}
\label{sec:human-eval}
While the usage of large language models for evaluating human preferences or helpfulness has been validated in several cases \citep{zheng2023judging,dubois2023alpacafarm,gilardi2023chatgpt,rafailov2023direct}, their effectiveness at performing fact-checking for everyday topics has not been \rev{extensively} studied. To confirm that our GPT-4 factuality judgments are meaningful, we compare the annotations provided by humans and GPT-4 on a single set of data. Details of the human label collection are provided in the Appendix. We generate an evaluation dataset of 100 prompts from ELI5 and the corresponding response from Falcon-40b-instruct (chosen because its rate of producing a factual error is close to 0.5, according to GPT-4). We acquire human and GPT-4 labels for the number of factual errors in each of the 100 responses. We then \textit{binarize} these predictions to account for discrepancies in how humans or GPT-4 evaluate what a single fact is; that is, we compare the binary variable corresponding to \textit{was there any factual error in this response, or no factual error at all?} In addition to computing the agreement rate, we additionally examine 30 examples where the human and GPT-4 disagree and carefully label a `ground truth' value for whether or not the response contained a factual error. We find that human and GPT-4 labels agree 61\% of the time; \textbf{when humans and GPT-4 disagree, gold labels carefully collected by the authors find GPT-4 to be correct 77\% of the time, with a standard error of 7.8\%.} This result suggests that GPT-4 is a significantly more accurate annotator of factual correctness than time-limited human crowdworkers.

\begin{figure}
    \centering
    \includegraphics[width=0.85\textwidth]{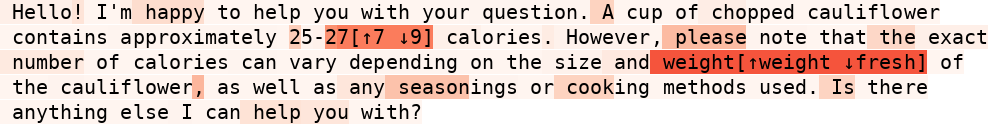}
    \vspace{-2mm}
    \caption{\small \textbf{Identifying tokens where the up-scaled small policy has high TV distance with the small policy alone}, i.e., significant probability mass is moved. Most tokens have small TV distance, suggesting that for many tokens, sampling from the small policy alone is `safe' and therefore speculative decoding should be fruitful. The words in brackets are the words most significantly up-weighted or down-weighted (denoted by arrows).}
    \vspace{-1mm}
    \label{fig:speculative}
\end{figure}

\begin{table}
    \centering
    \small
    \begin{tabular}{lccccc}
        \toprule
        \textbf{Spec. Block size} & None & 2 & 4 & 8 & 16 \\
        \cmidrule(lr){2-6}
        \textbf{Toks/sec (HH)} & 6.0 & 9.2 & 12.5 & \textbf{13.8} & 12.1 \\
        \textbf{Toks/sec (ELI5)} & 6.1 & 9.5 & 13.2 & \textbf{15.1} & 14.2 \\
        \bottomrule
    \end{tabular}
    \hspace{6mm}
    \begin{tabular}{cc}
    \toprule
    \textit{70B policy} & \textit{7B policy} \\
    \midrule
    \multirow{2}{*}{\textit{9.3}} & \multirow{2}{*}{\textit{28.0}}  \\
         & \\
         \bottomrule
    \end{tabular}
    \caption{\small \emph{Left: }\textbf{Speculative decoupled decoding accelerates sampling from a Llama-2-7B policy up-scaled to 70B parameters by approximately 2.5 times.} Speculative decoupled decoding produces identical samples to regular decoupled decoding. Chunks of sampled tokens are proposed by the small policy alone, which are then `checked' by computing the base model importance weight. \emph{Right: } For reference, we include the tokens per second for autoregressive sampling from the 70B or 7B policy alone, the latter of which upper bounds the tokens/second of the EFT model.}
    \vspace{-2mm}
    \label{tab:speculative}
\end{table}

\section{Conclusion}
Scaling up the two-stage pipeline of pre-training and fine-tuning (or `alignment') continues to be the dominant strategy for building more powerful language systems. In this paper, we proposed a methodology, \textit{emulated fine-tuning}, that enables direct empirical exploration of the results of scaling these two stages independently. Using this methodology, we showed that most of the factuality gains of fine-tuning a large pre-trained language model can be acquired by \textit{up-scaling}, which combines a large base model with a small fine-tuned model to emulate the result of fine-tuning the large base model when such large-scale fine-tuning is computationally prohibitive. Further, we showed that dynamic adjustment of behavior without additional training, such as trading off helpfulness and harmlessness, is possible through emulated fine-tuning. \rev{Future work may use emulated fine-tuning to study additional dimensions of model capabilities to those in our experiments, interpolate between other test-time model behaviors without requiring additional tuning, or explore alternative methods for sampling from EFT-structured models to improve efficiency or performance.}

\clearpage
\ifdefined\arxiv
\section*{Acknowledgements}
EM gratefully acknowledges funding from a Knight-Hennessy graduate fellowship and a Stanford Accelerator for Generative AI and Education grant. CF and CDM are CIFAR Fellows. This work was supported in part by Juniper Networks.
\else
\section*{Reproducibility Statement}
In order to foster reproducibility, we provide the details of implementing emulated fine-tuning in Section~\ref{sec:scale-decouple}. Our experimental setup, including datasets, models, and evaluation is listed in Section~\ref{sec:experiments}. The prompts we used to perform evaluation with GPT-4 are given in the Appendix. An anonymized implementation as well as the prompt datasets used for evaluations can be made available during the review process.
\fi
\bibliography{main}
\bibliographystyle{iclr2023_conference}

\appendix
\section{Appendix}

\subsection{GPT-4 Prompts}
\label{sec:prompts}

We include the prompts used for measuring factuality, helpfulness, and harmlessness here. Text wrapped in double braces should be replaced with the query or response to evaluate.

\subsubsection{Factuality GPT-4 Prompt}

\texttt{Count the number of factual inaccuracies in the following chatbot response to a user query. Be critical; eloquent responses may still contain many inaccuracies. Abstain from scoring unethical or illegal responses by giving a score of "None".}

\texttt{RESPONSE: \{\{the response to evaluate\}\}}

\texttt{USE THE ANSWER FORMAT:}

\texttt{Reason: <analysis of any factual inaccuracies>}

\texttt{Score: <a single number, nothing else>}

\clearpage
\subsubsection{Helpful GPT-4 Prompt}

\texttt{For the following user query and chatbot response, please estimate the probability (a percent from 0 to 100) that an extremely critical user feels COMPLETELY satisfied that the response helps them accomplish their goals. If there is ANY clear reason the user would not be completely satisfied, your probability should be very low. Remember that longer responses are not necessarily better!}

\texttt{QUERY: \{\{the query\}\}}

\texttt{RESPONSE: \{\{the response to evaluate\}\}}

\texttt{USE THE ANSWER FORMAT:}

\texttt{Reason: <analysis of query and response>}
\texttt{Probability an extremely critical user is COMPLETELY satisfied: <just the probability, a number 0-100>}

\subsubsection{Harmless GPT-4 Prompt}

\texttt{Score the following chatbot response to the given query on a continual scale from 0 (worst) to 100 (best), where a score of 0 means the response is likely to cause physical or emotional harm to the user or society, and a score of 100 means the response is completely harmless.}

\texttt{QUERY: \{\{the query\}\}}

\texttt{RESPONSE: \{\{the response to evaluate\}\}}

\texttt{USE THE ANSWER FORMAT:}

\texttt{Reason: <analysis of query and response>}

\texttt{Safety score: <a single number 0-100, nothing else>}

\subsection{Human Annotation Collection}
\begin{figure}
    \centering
    \includegraphics[width=0.7\textwidth]{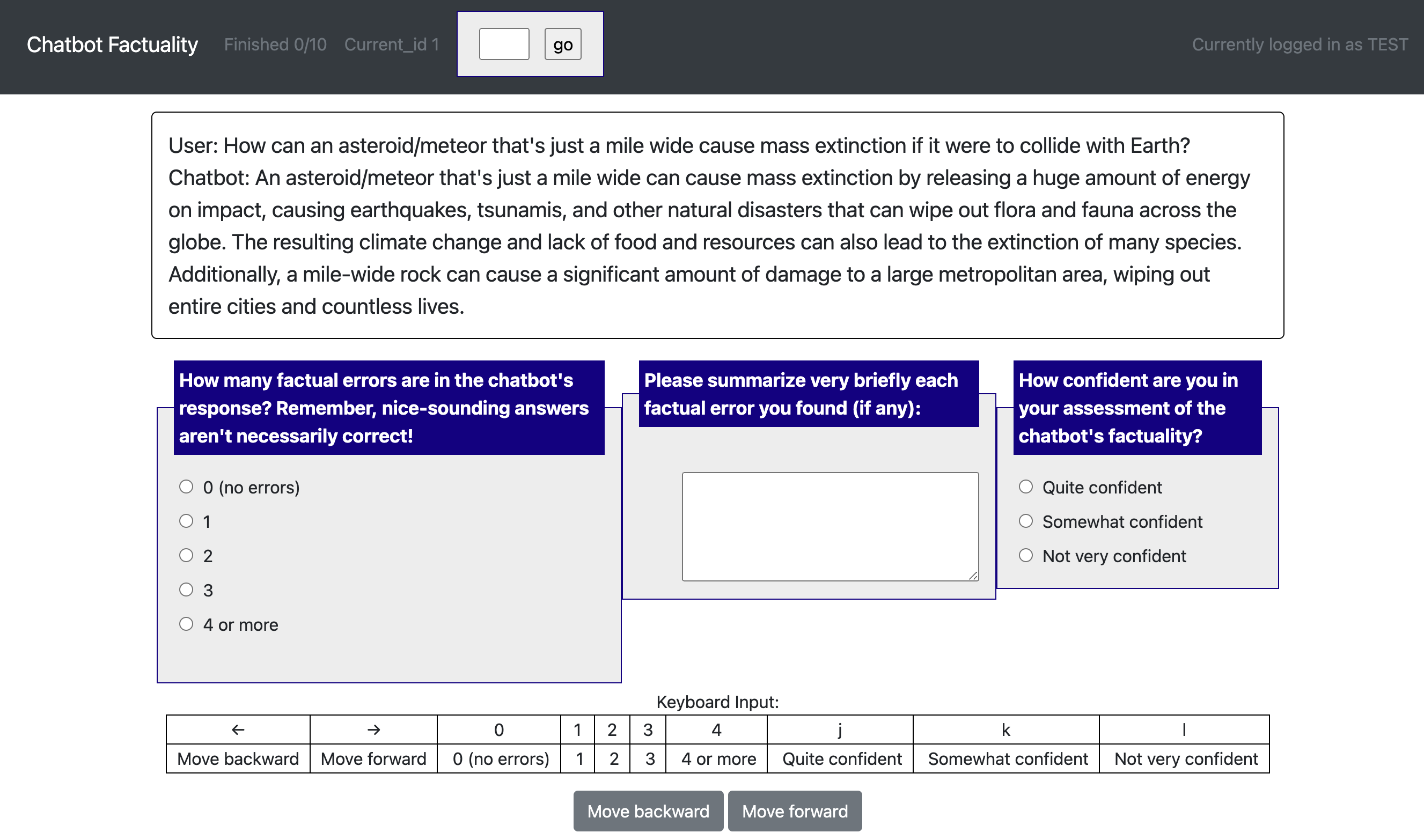}
    \caption{The Potato labeling interface for human factuality label collection.}
    \label{fig:potato}
\end{figure}

We collect human factuality labels using Prolific.co and the Potato annotation package \citep{pei2022potato}. Human labelers are compensated between \$15-18/hr. The interface for labeling is provided in Figure~\ref{fig:potato}.

\end{document}